\documentclass[conference]{IEEEtran}
\usepackage{times}

\usepackage{multicol}
\usepackage[bookmarks=true]{hyperref}
\usepackage{multirow}
\usepackage{array}
\usepackage{hyperref}
\usepackage{url}
\usepackage{float}
\usepackage{tabularx}  
\usepackage{booktabs}
\usepackage{xcolor} 
\usepackage{colortbl}
\usepackage{mathtools}
\usepackage{amsfonts}  
\usepackage{graphics}
\usepackage[pdftex]{graphicx}
\usepackage{wrapfig}
\usepackage{caption}
\usepackage{color}
\usepackage{dsfont}
\usepackage[]{mdframed}
\usepackage{float}
\usepackage[ruled,vlined,linesnumbered]{algorithm2e}
\usepackage{xparse}
\usepackage{amsmath}
\usepackage{bm}
\usepackage{mathtools}
\usepackage{amssymb}
\usepackage{amsthm}
\usepackage{amssymb}

\usepackage{booktabs}
\usepackage{epigraph}

\usepackage{xcolor} 
\usepackage{hyperref} 
\hypersetup{
    colorlinks=true,
}

\usepackage{xargs} 
\usepackage[colorinlistoftodos,prependcaption,textsize=tiny]{todonotes}
\newcommandx{\wrn}[2][1=]{\todo[linecolor=red,backgroundcolor=red!25,bordercolor=red,#1]{#2}}
\newcommandx{\cmt}[2][1=]{\todo[linecolor=blue,backgroundcolor=blue!25,bordercolor=blue,#1]{#2}}

\IEEEoverridecommandlockouts

\begin{document}
\makeatletter
\let\@oldmaketitle\@maketitle
\renewcommand{\@maketitle}{\@oldmaketitle
  \begin{center}
  \captionsetup{type=figure}
  \setcounter{figure}{0}
  \includegraphics[width=1.0\textwidth,height=0.32\textheight]{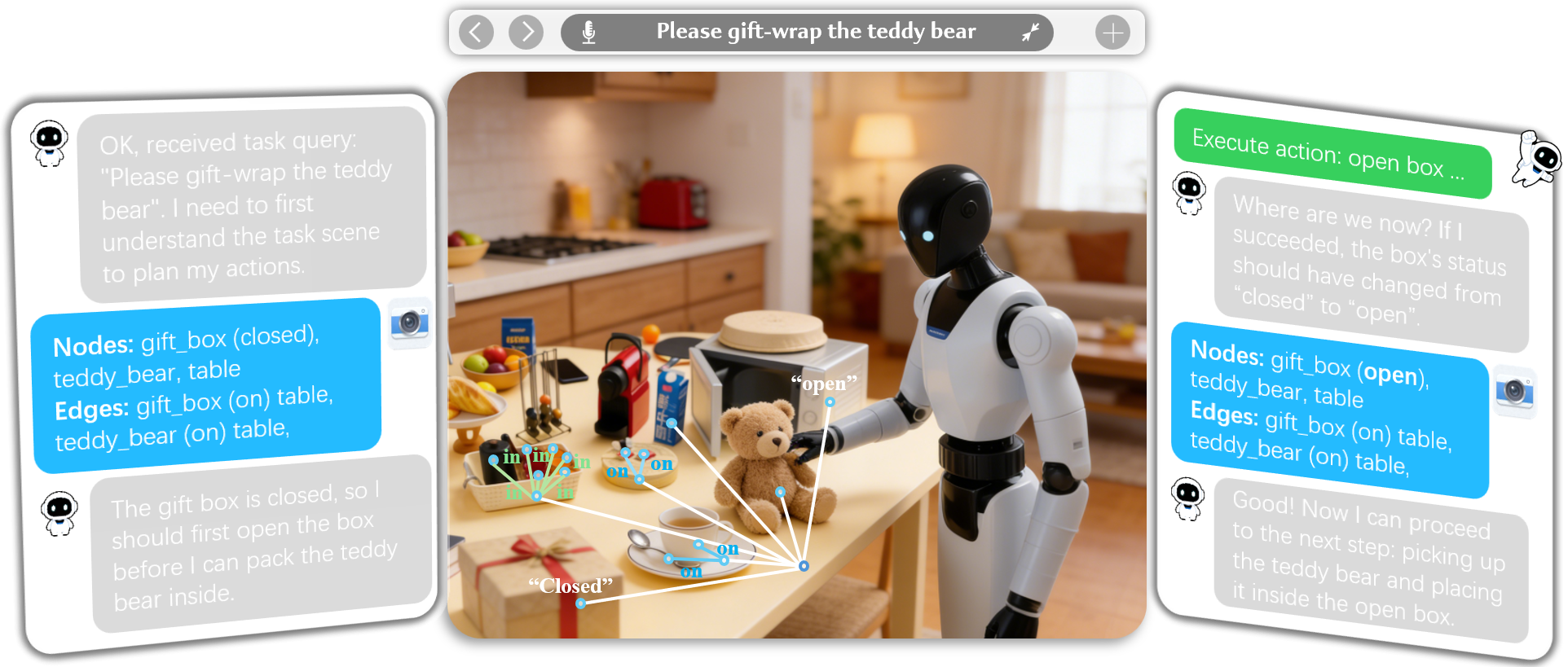}
    \captionof{figure}{RoBoSR constructs a structured scene representation that explicitly models objects, spatial relations, and affordances. This structured representation serves as an intermediate reasoning substrate, enabling robust task reasoning and planning} 
    \label{fig:overall of data}
    \end{center}
}

\title{RoBoSR: Structured Scene Representations for Embodied Robotic Reasoning}



\def\cameraready{0}  

\ifx\cameraready\undefined
    \author{
    Anonymous Submission
    }
\else
    \author{
    Kewei Hu,
    Wanchan Yu,
    Fangwen Chen,
    Jiajian Jing,
    Zimeng Li,
    Wei Ying,\\
    Tianhao Liu,
    Michael Zhang,
    Hanwen Kang$^{*}$
    }
\fi

\maketitle
 
\begin{abstract}
Despite rapid progress, embodied reasoning under real-world variability remains challenging. Existing approaches rely on demonstration-driven sequential biases, limiting flexibility in open-ended and long-horizon tasks that require structured reasoning over evolving states.
We introduce RoBoSR, an intermediate structural representation that formulates manipulation as step-wise state transitions over semantically grounded, object-centric scene graphs. By modeling object states and their spatial relations at the perception–action interface, RoBoSR disentangles high-level task reasoning from raw inputs and enables structured reasoning over preconditions, effects, and goal states. This representation endows the agent with causal reasoning capability, enforcing subtask dependencies and supporting coherent long-horizon task planning.
To learn such structure-aware reasoning, we construct Manip-Cognition-1.6M, an open-world dataset that jointly supervises scene understanding, instruction interpretation, and subtask planning across diverse tasks.
Across several benchmarks, and real-world demonstration, our method consistently outperforms prompting-based methods and classical TAMP baselines in zero-shot generalization and long-horizon task. The results underscore structured intermediate representations as a critical inductive bias for scalable embodied reasoning.
\end{abstract}

\IEEEpeerreviewmaketitle

\section{Introduction}
\IEEEPARstart{E}{mbodied} learning has become a promising paradigm for building robotic systems that integrate perception, reasoning, planning, and execution. 
Despite recent progress, enabling robust task reasoning under real-world variability remains challenging \cite{black2024pi_0}. 
Many existing approaches rely heavily on human demonstrations, where tasks are performed through relatively consistent procedural sequences. 
While effective within the training distribution, such sequential inductive biases may limit flexibility when agents encounter open-ended tasks with multiple valid execution paths or long-horizon tasks that require reasoning over extended dependencies \cite{zhao2025cot}.

An embodied agent relies on two fundamental capabilities: high-level task reasoning and low-level action control. 
Embodied reasoning infers intention-oriented action sequences (e.g., “\textit{open the refrigerator}” or “\textit{place the cup on the plate}”) from perceptual inputs \cite{chen2024roboscript}, requiring the agent to interpret the environment, understand task intent, and generate coherent action plans. 
Central to this process is how the task scene is represented. A task scene representation must capture object-centric entities, spatial relationships, and task-relevant semantics, as different goals within the same scene can induce distinct action sequences, and conversely, changes in scene configuration may alter the required behaviors under the same goal.
Existing learning-based approaches typically encode such information implicitly within high-dimensional neural features. While effective for end-to-end policy learning, these representations do not explicitly model relational structure, which can limit systematic reasoning and generalization across scenes and tasks. 
On the other hand, classical Task and Motion Planning (TAMP) frameworks provide structured symbolic representations (e.g., PDDL or BDDL), enabling explicit reasoning, but they often impose complex and rigid format constraints, requiring manually predefined abstractions that are difficult to scale in open-ended, dynamic environments \cite{ji2025robobrain}.
These observations suggest a need for structured intermediate representations that combine the flexibility of neural perception with the explicit relational structure required for robust embodied reasoning.
Motivated by these limitations, this study explores: \textbf{Can embodiments leverage structured intermediate representations to enhance reasoning abilities general tasks?}

In this study, we introduce \textbf{RoBo}tic \textbf{S}cene \textbf{R}epresentation (\textbf{RoBoSR}), an embodied intermediate representation that formulates decision making as reasoning over structured scene states. 
RoBoSR leverages semantically grounded scene graphs to extract object-level and relational structure from observations, explicitly separating high-level state reasoning from low-level action execution. 
To train agent over \textbf{RoBoSR}, we further construct the \textbf{Manip-Cognition-1.6M}, which provides joint supervision over \textit{scene understanding}, \textit{instruction interpretation}, and \textit{sub-task planning} across a range of tasks.
The agent reasoning on \textbf{RoBoSR} introduces three key capabilities.
\begin{itemize}
    \item \textbf{Instruction–State Alignment}: By grounding reasoning in an object-centric representation, the agent maps high-level instructions to scene-consistent actions, supporting instruction following in open-world tasks.
    \item \textbf{Causally Constrained State-Space Reasoning}: RoBoSR formulates reasoning as explicit state evolution, constraining subtask planning through causal structure rather than fixed demonstration sequences, thereby supporting coherent long-horizon reasoning in open-ended tasks.
    \item \textbf{State-consistency correction}: The explicit and comparable state representation in RoBoSR makes execution deviations observable, enabling correction of inconsistencies through structured reasoning.  
\end{itemize}


\section{Related Works}
Scene representation plays a central role in embodied decision making, as it determines what structure is available for reasoning and generalization. 
Existing approaches largely fall into two paradigms. 
Classical symbolic frameworks, such as PDDL-based planning and Task and Motion Planning (TAMP), model the world as explicitly defined symbolic states with manually specified predicates and transition rules. 
This formulation enables compositional reasoning and formal search over structured state spaces, but typically relies on predefined vocabularies and engineered domain knowledge, limiting adaptability under perceptual uncertainty and open-world variability. 
In contrast, recent end-to-end Vision-Language-Action (VLA) models, including RT-2~\cite{zitkovich2023rt}, OpenVLA~\cite{kim2024openvla}, and $\pi_0$~\cite{black2024pi_0}, directly map high-dimensional visual observations to actions through large-scale latent representations. 
These approaches benefit from data scaling and avoid manual symbolic specification. 
However, operating primarily in pixel or latent embedding space introduces a representation bottleneck~\cite{ai2025review}, in which task-irrelevant visual variability can become entangled with task-relevant physical relations. 
Such entanglement may reduce robustness in long-horizon manipulation and under distribution shift~\cite{brohan2022rt}. 
Hierarchical variants, such as RT-H~\cite{belkhale2024rt}, introduce planning–control decomposition, yet their intermediate representations often remain implicit and lack explicit supervision over object states and spatial relations~\cite{yang2023foundation}. 
These observations suggest that architectural decomposition alone is insufficient; the inductive bias imposed by the chosen state representation fundamentally shapes reasoning capability.

Scene graphs provide a structured yet learnable intermediate representation by explicitly modeling objects, attributes, and spatial relations~\cite{li2024scene}. 
In robotics, scene graphs have been employed for perception, mapping, and language-conditioned planning support~\cite{gu2024conceptgraphs, wang2025enact}. 
Works such as VoxPoser~\cite{huang2023voxposer} and ReKep~\cite{huang2024rekep} translate language instructions into motion constraints or relational keypoints, while ConceptGraphs~\cite{gu2024conceptgraphs} and ENACT~\cite{wang2025enact} leverage graph structures to improve spatial understanding. 
Large language models have also demonstrated the ability to reason over graph-structured inputs. 
These developments highlight the compatibility between object-centric graph representations and language-based reasoning. 
However, in most prior systems, scene graphs function primarily as perceptual abstractions or auxiliary planning signals; decision making itself remains prompt-driven or does not explicitly model state transitions in graph space. 
Consequently, the scene graph does not fully serve as the core state variable governing embodied reasoning dynamics.

\begin{figure*}
    \centering
    \includegraphics[width=1.0\textwidth,height=0.275\textheight]{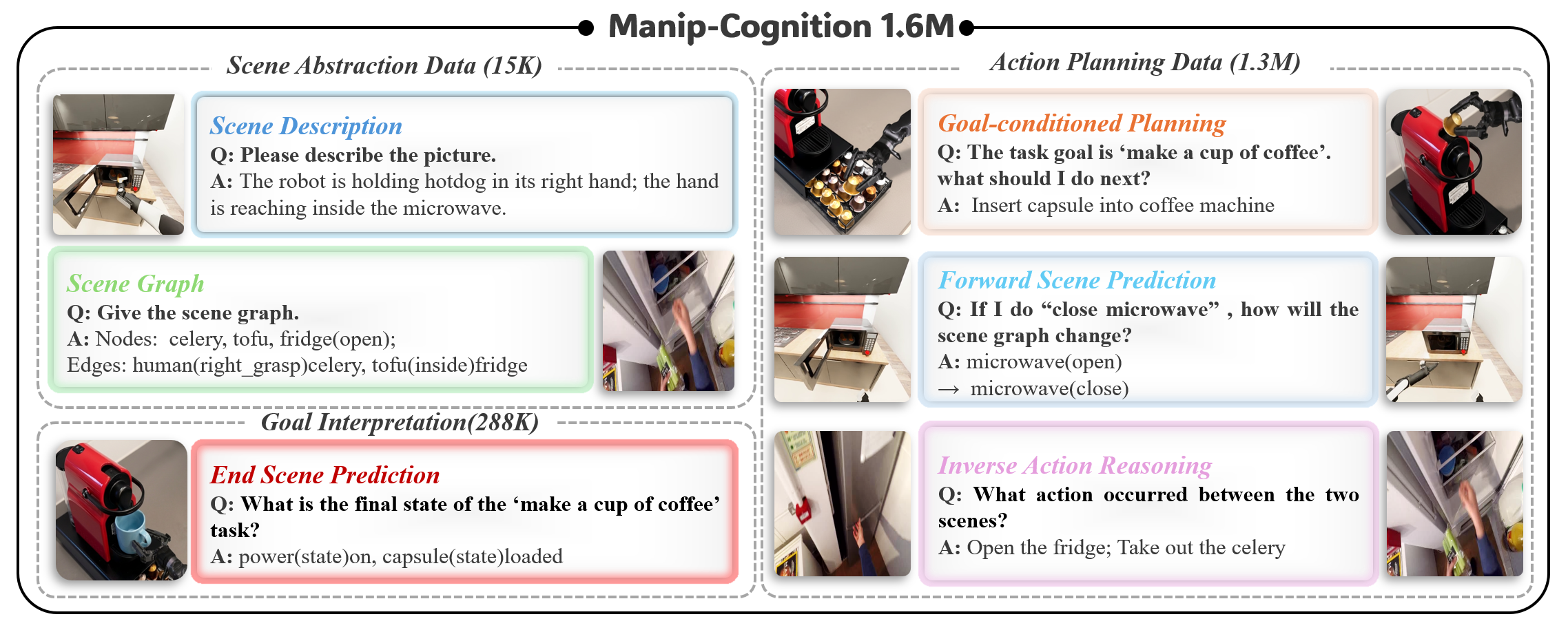}
    \caption{Examples of the Manip-Cognition data including scene abstraction data, planning data and goal interpretation data}
    \label{fig:dataset}
\end{figure*}

RoBoSR adopt a different formulation: the grounded scene graph is treated as the primary state space for decision making, and reasoning is explicitly modeled as state transitions within this structured space. 
Unlike classical symbolic planners, both the representation and transition reasoning are learned from large-scale egocentric data rather than predefined rules. 
Unlike latent end-to-end VLAs, action selection operates over object-centric states with explicit relational predicates. 
This design introduces an inductive bias that separates relational task structure from perceptual variability, while preserving adaptability through data-driven learning. 
By jointly learning goal-state inference, forward state evolution, and next-action prediction, RoBoSR enables step-wise reasoning over states and integrates naturally with perception and control pipelines.

\section{Methodology} \label{sec: method}

\subsection{RoBoSR Modeling}
Given an RGB-D observation $\mathcal{I}$, we construct RoBoSR, a structured 3D scene graph $M_{sg} = (O_t, E_t)$ to represent the workspace and object states from multiple affordance perspectives. Figure \ref{fig:dataset} illustrates the transformation process from raw visual input to a structured scene graph.  

\textbf{Object affordance:} Each object $o_j \in O_t$ is represented as a structured entity capturing its functional components, articulated child parts, and open/closed states when applicable. For example, a \textbf{mug} includes a functional keypoint corresponding to its \textbf{handle}, while a \textbf{cabinet} is modeled as an articulated object with multiple child elements, such as \textbf{drawers}, whose states (open/closed) are explicitly represented.  

\textbf{Spatial affordance:} Each edge $e_k \in E_t$ encodes a spatial relationship between a pair of objects, representing functional and geometric interactions in the environment. These relations include canonical predicates such as \textit{on}, \textit{inside}, \textit{adjacent}, as well as more nuanced configurations like \textit{tilted against} or \textit{partially occluded}. For example, a \textbf{mug} may be \textit{on} a \textbf{table}, whereas a partially opened \textbf{drawer} is \textit{adjacent} to the cabinet body, reflecting both spatial proximity and kinematic state.
 
By combining these affordances, $M_{sg}$ provides a structured, functionally grounded, and relationally rich representation suitable for reasoning and planning in manipulation tasks.

\subsection{RoBoSR Data Engine}
We introduce \textbf{Manip-Cognition-1.6M}, a structured dataset designed to support scene abstraction, action planning, and goal interpretation (see Figure.~\ref{fig:dataset}). 
The data engine provides supervision for sequential reasoning in open-ended, long-horizon, instruction-driven tasks.
The dataset is constructed from Epic-Kitchens-100~\cite{damen2020epic}, EgoPlan~\cite{chen2023egoplan}, Behavior-1K~\cite{li2024behavior}, and Enact~\cite{wang2025enact}. 
In total, we collect 6k task trajectories spanning daily activities (e.g., kitchen manipulation, tabletop organization, object rearrangement, and pick-and-place scenarios).
Through step-wise decomposition and augmentation, the final dataset contains 1.6M samples.

\subsubsection{\textbf{Scene Abstraction Data (15k)}}
This data subset comprises 15k annotated pairs after linguistic and structural augmentation. 
It trains the model to construct structured scene graphs ($SG$) from natural language. 
The input consists of language descriptions of object configures and spatial relations, and the output is semantically grounded scene graph:
\[
\text{Text}_{t} \leftrightarrow  SG_{t}.
\]
The supervision targets two capabilities: (1) object grounding, i.e., \textit{identifying valid object entities}, and (2) spatial relation extraction, i.e., \textit{inferring relational edges between objects} in structured form.

\subsubsection{\textbf{Action Planning Data (1.3M)}}
This data subset comprises 1.3M samples derived from 6k task trajectories. 
It trains the model to capture causal dependencies between objects, actions, and structured world states for sequential decision-making.
We decompose planning supervision into three complementary objectives:

(a) Goal-conditioned Planning. 
The model predicts the immediate next symbolic action $A_t$ given the current scene graph $SG_t$ and a high-level task goal $G$:
\begin{equation}
    (SG_t, G) \rightarrow A_t
\end{equation}
This objective trains goal-directed action selection conditioned on structured state representations.

(b) Forward Scene Prediction.
The model predicts the structural state transition $\Delta SG_{\text{edge}}$ induced by an action:
\begin{equation}
    (SG_t, A_t) \rightarrow \Delta SG_{\text{edge}}
\end{equation}
where $\Delta SG_{\text{edge}}$ denotes edge changes between $SG_t$ and $SG_{t+1}$. 
This objective supervises modeling of action-induced environmental dynamics.

(c) Inverse Action Reasoning.  
Given an initial state $SG_t$ and a target state $SG_{t+n}$, the model infers the multi-step action sequence required to achieve the transition:
\begin{equation}
    (SG_t, SG_{t+n}) \rightarrow \{A_i\}_{i=t}^{t+n-1}
\end{equation}
This objective trains multi-step causal reasoning in structured space.

\subsubsection{\textbf{Goal Interpretation Data (0.3M)}}
This data subset comprises 0.3M samples. 
It trains the model to infer structured goal states from partial world configurations and natural language task instructions.
Given a current scene graph $SG_t$ and a high-level textual instruction, the model predicts the corresponding target scene graph $SG_{goal}$:
\begin{equation}
    (SG_t, \text{Instruction}) \rightarrow SG_{goal}
\end{equation}
Here, $SG_{goal}$ encodes the desired object states and spatial relations that characterize successful task completion. 
Unlike step-wise transition modeling, this objective supervises direct goal-state abstraction in structured space, requiring the model to reason over long-horizon task semantics and implicit intermediate steps.
For each trajectory, we pair intermediate scene graphs with the final state defined by the instruction, enabling supervision over instruction-conditioned goal inference.

\subsection{Training Agentic Reasoning over RoBoSR}

RoBoSR is instantiated using the Qwen3-8B architecture \cite{qwen8B_system_card_2025}. 
At inference time, RoBoSR takes as input the scene graph and a user instruction, and autoregressively predicts the next action command. 
Training proceeds in two stages: (1) SFT to induce structured reasoning over RoBoSR, followed by (2) RFT to enforce consistency with execution constraints.

\subsubsection{\textbf{Supervised Fine-Tuning}}
The first stage learns structured scene interpretation, goal grounding, and action reasoning via supervised fine-tuning. 
We train on instruction-following trajectories conditioned on RoBoSR representations, using Low-Rank Adaptation applied to all linear layers for parameter-efficient adaptation.
The resulting policy $\pi_{\mathrm{SFT}}$ generates intermediate reasoning traces followed by executable symbolic actions. 
However, supervised training alone may yield object-grounding errors and inconsistent state transitions, motivating a second-stage reinforcement-based alignment procedure.

\begin{table}[ht]
\centering
\caption{Training Hyperparameters for RoBoSR Models}
\label{tab:hyperparams}
\renewcommand{\arraystretch}{0.85}
\begin{tabular}{@{}lc@{}}
\toprule
\textbf{Hyperparameter} & \textbf{Qwen3-8B} \\ \midrule
Model Scale & 8.2 B \\
Non-Embedding Parameters & 6.95 B \\
Max Context Length & 131,072 (YaRN) \\
Training Sequence Length & 2,048 \\
Precision & bfloat16 \\
Training Method & LoRA (all-linear) \\
LoRA Rank ($r$) & 8 \\
LoRA Alpha ($\alpha$) & 32 \\ \midrule
Optimizer & AdamW \\
Learning Rate & $1 \times 10^{-4}$ \\
LR Scheduler & Cosine w/ Warmup \\
Warmup Ratio & 0.05 \\
Training Epochs & 20 \\
Per-Device Batch Size & 4 \\
Gradient Accumulation Steps & 16 \\
Global Batch Size & 128 \\ \bottomrule
\end{tabular}
\end{table}

\subsubsection{\textbf{Reinforcement Fine-Tuning}}
We perform reinforcement fine-tuning using Group Relative Policy Optimization (GRPO). 
The objective maximizes the following task-level reward:
\begin{equation}
R_{\text{total}} = \lambda_1 R_{S} + \lambda_2 R_{G} + \lambda_3 R_{T}.
\end{equation}
where $R_{S}$ penalizes multi-step action hallucinations, $R_{G}$ enforces object-grounding consistency with the current scene graph, and $R_{T}$ regulates proper task termination. 
The coefficients $\lambda_1$, $\lambda_2$, and $\lambda_3$ balance the contributions of the three reward components, respectively.

\paragraph{Step-wise Constraint Reward}
This reward term enforces strict step-wise action generation and prevents multi-action hallucinations (e.g., producing multiple primitives in a single response). 
Given the predicted action sequence $A_{\text{pred}}$, the reward $R_S$ penalizes outputs containing more than one executable primitive and only permits a single atomic action following the reasoning trace:
\begin{equation}
R_{S} =
\begin{cases}
1,      & \text{if } a_{\text{pred}}=\texttt{END}\land \text{Satisfied}(S,G)\\[2pt]
-\alpha, & \text{if } N(A_{\text{pred}})>1\\[2pt]
0,      & \text{otherwise}
\end{cases}
\end{equation}
where $N(\cdot)$ counts executable action primitives. This constraint enforces closed-loop, step-wise reasoning.

\paragraph{Scene-Graph Grounding Reward}

This term penalizes object-grounding errors by enforcing consistency with the current RoBoSR graph. 
Given the predicted object identifier $obj_{\text{pred}}$ and the set of valid node labels $V_{SG}$, the reward is defined as
\begin{equation}
R_{G} = \mathbb{I}(obj_{\text{pred}} \in V_{SG}),
\end{equation}
where $\mathbb{I}(\cdot)$ denotes the indicator function. This eliminates invalid object references during execution.

\paragraph{Termination Reward}

This term regulates stopping behavior to prevent premature termination or infinite action generation. 
\begin{equation}
R_{T} =
\begin{cases}
1,  & \text{if } a_{\text{pred}}=\texttt{END}\ \land\ \text{Satisfied}(S_{\text{env}},G) \\[4pt]
-\beta, & \text{if } a_{\text{pred}}=\texttt{END}\ \land\ \neg\,\text{Satisfied}(S_{\text{env}},G) \\[4pt]
0,  & \text{otherwise}
\end{cases}
\end{equation}
where $S_{\text{env}}$ denotes the current environment state and $G$ the goal condition. This reward differentiates correct termination from erroneous stopping.

\subsection{Physical System Deployment} \label{sec: real_robot}
To deploy RoBoSR on a physical platform, we implement a modular perception–reasoning–action architecture. 
The system comprises a perception–reasoning module for structured decision-making and an action expert for low-level control execution.
For perception, we adopt an open-world segmentation network based on SAM3 to extract object-centric regions from RGB-D observations. 
A geometry-conditioned rule-based algorithm then constructs RoBoSR by obtaining object states and spatial relations from the segmented outputs. 
The resulting structured world representation is provided to RoBoSR for step-wise reasoning and action selection.
For action execution, we employ Semantic Keypoint Imitation Learning (SKIL) \cite{wang2025skil} as the low-level controller. 
Given the instance regions identified by SAM3, masked observations corresponding to the selected object are passed to SKIL to generate executable motion trajectories. 
For each manipulation primitive involved in our task set (e.g., \textit{"grasp a mug handle"}), we collect 50 demonstrations and train a dedicated SKIL policy.
\begin{table*}[!htbp]
\centering
\caption{Task Progress (TP) comparison across different methods on three robotic manipulation tasks. Best results are highlighted in \textbf{bold} and our method is shaded in gray.}
\label{tab:task_progress}
\setlength{\tabcolsep}{4pt}  
\resizebox{\textwidth}{!}{%
\renewcommand{\arraystretch}{0.5}
\footnotesize 
\begin{tabular}{lccccccccc}
\toprule
\multirow{2}{*}{Method} & \multicolumn{3}{c}{Semantic Object Disambiguation} & \multicolumn{3}{c}{Spatial-Aware Sequencing} & \multicolumn{3}{c}{Goal-conditioned Generalization} \\
\cmidrule(lr){2-4} \cmidrule(lr){5-7} \cmidrule(lr){8-10}
 & Easy & Medium & Difficult & Easy & Medium & Difficult & Easy & Medium & Difficult \\
\midrule
Gemini-2.5-pro & 0.323 & 0.278 & 0.175 & 0.230 & 0.170 & 0.090 & 0.234 & 0.050 & 0.020 \\
GPT-5 & 0.300 & 0.280 & 0.100 & 0.110 & 0.090 & 0.020 & 0.130 & 0.050 & 0.010 \\
Claude-sonnet-4.5 & 0.280 & 0.280 & 0.190 & 0.150 & 0.080 & 0.020 & 0.200 & 0.060 & 0.000 \\
DeepSeek-V3 & 0.230 & 0.160 & 0.150 & 0.130 & 0.120 & 0.070 & 0.160 & 0.090 & 0.040 \\
Qwen3-8B & 0.630 & 0.550 & 0.410 & 0.064 & 0.000 & 0.000 & 0.130 & 0.067 & 0.017 \\
RoBoSR-8B & 0.700 & 0.610 & 0.520 & 0.330 & 0.230 & 0.150 & 0.300 & 0.190 & 0.080 \\
\rowcolor{gray!20}
RoBoSR-8B-FT & \textbf{0.975} & \textbf{0.917} & \textbf{0.892} & \textbf{0.934} & \textbf{0.812} & \textbf{0.762} & \textbf{0.891} & \textbf{0.291} & \textbf{0.131} \\
\bottomrule
\end{tabular}%
}
\end{table*}





\section{Experiments} \label{sec: exp}
We conduct a comprehensive evaluation of RoBoSR, including spatial-aware, goal-conditioned, and long-horizon reasoning in the \textbf{GSR-bench} \cite{GSRbench_2026}, and real-world robot implementation to assess performance. Unless otherwise specified, \textbf{RoBoSR-8B} denotes the base model trained solely on the Manip-Cognition-1.6M dataset without exposure to any GSR-bench tasks.
\textbf{RoBoSR-8B-FT} further applies a lightweight post-training stage on top of RoBoSR-8B using 36 in-domain demonstration trajectories from GSR-bench.

\subsection{Evaluation on GSR-bench}
\label{sec:gsrbench_eval}
\textbf{GSR-bench} evaluates long-horizon task reasoning under both spatial and semantic constraints. 
It consists of 180 tasks, with an average planning horizon exceeding 10 steps. 
Our evaluation focuses on three aspects: 
\textbf{(1) Semantic Object Disambiguation (SOD)}, which assesses reasoning under varying object semantics; 
\textbf{(2) Spatial-Aware Sequencing (SAS)}, which evaluates reasoning over physical causality and spatial constraints; 
and \textbf{(3) Goal-conditioned Generalization (GCG)}, which measures reasoning across diverse abstract goals. 
For each aspect, we define three difficulty levels: \textit{easy}, \textit{medium}, and \textit{difficult}. 
We use \textbf{Task Progress (TP)} as the evaluation metric, and the experimental results are reported in Table.\ref{tab:task_progress}.
Based on GSR-bench evaluation, we draw the following key conclusions:

 \begin{figure}[ht]
    \centering
    \includegraphics[width=1\linewidth]{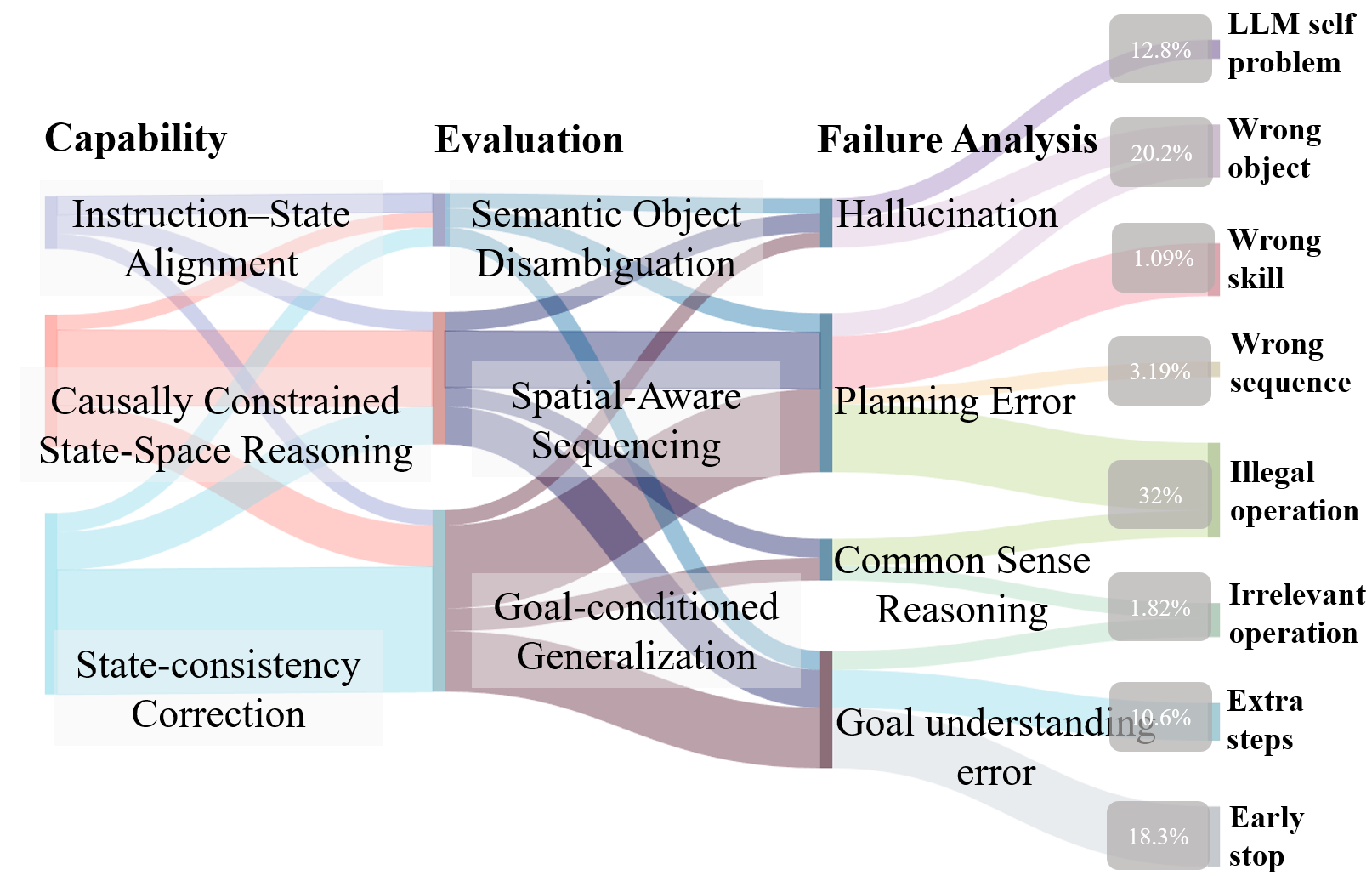}
    \caption{Failure Mode Attribution and Capability Decomposition on GSR-bench}
    \label{fig:Failure Mode}
\end{figure}

\textbf{First, structured scene-centric reasoning is a necessary inductive bias for long-horizon embodied tasks.}
Across all difficulty levels, RoBoSR-8B consistently outperforms prompting-based LLM baselines by a large margin, particularly on SAS and GCG.
While large language models exhibit rapid performance degradation as task complexity increases, RoBoSR maintains stable task progress by explicitly reasoning over object states and spatial relations.

\textbf{Second, large-scale structured pretraining enables strong zero-shot generalization, but task grounding remains a bottleneck.}
RoBoSR-8B, trained solely on Manip-Cognition-1.6M without exposure to GSR-bench tasks, demonstrates robust zero-shot reasoning.
However, its performance gap between easy and difficult settings indicates that abstract task semantics and termination conditions are not fully resolved by pretraining alone.

\textbf{Third, minimal post-training yields a promising performance improvement, highlighting the data efficiency of structured representations.}
With only 36 in-domain demonstration trajectories, RoBoSR-8B-FT achieves  over 60\% task completion rates on easy and medium tasks and substantially improves difficult goal-conditioned reasoning.
This improvement suggests that RoBoSR provides a highly sample-efficient substrate for adaptation, where a small amount of task-specific supervision suffices to align high-level reasoning with environment-specific constraints.

\subsection{Ablation Studies on GSR-bench}
\label{sec:gsrbench_ablation}

We conduct comprehensive ablation studies on GSR-bench to validate the contribution of each core component in our method. All ablated models are initialized from the \textbf{RoBoSR-8B} checkpoint, and \textbf{Full model} (i.e., \textbf{RoBoSR-8B-FT}) serves as the primary baseline, representing the complete system after both pre-training and task-specific post-training. We systematically remove or alter individual components from this full model to isolate their effects.

We evaluate performance using Task Progress (TP) and categorize failures into Hallucination Error (HE), Planning Error (PE), Common Sense Reasoning error (CSR), and Goal Understanding Error (GUE). 
Figure.~\ref{fig:Failure Mode} visualizes the failure-mode distribution of the full model and illustrates how different reasoning demands manifest across task categories.

\begin{figure}[ht]
    \centering
    \includegraphics[width=0.75\linewidth]{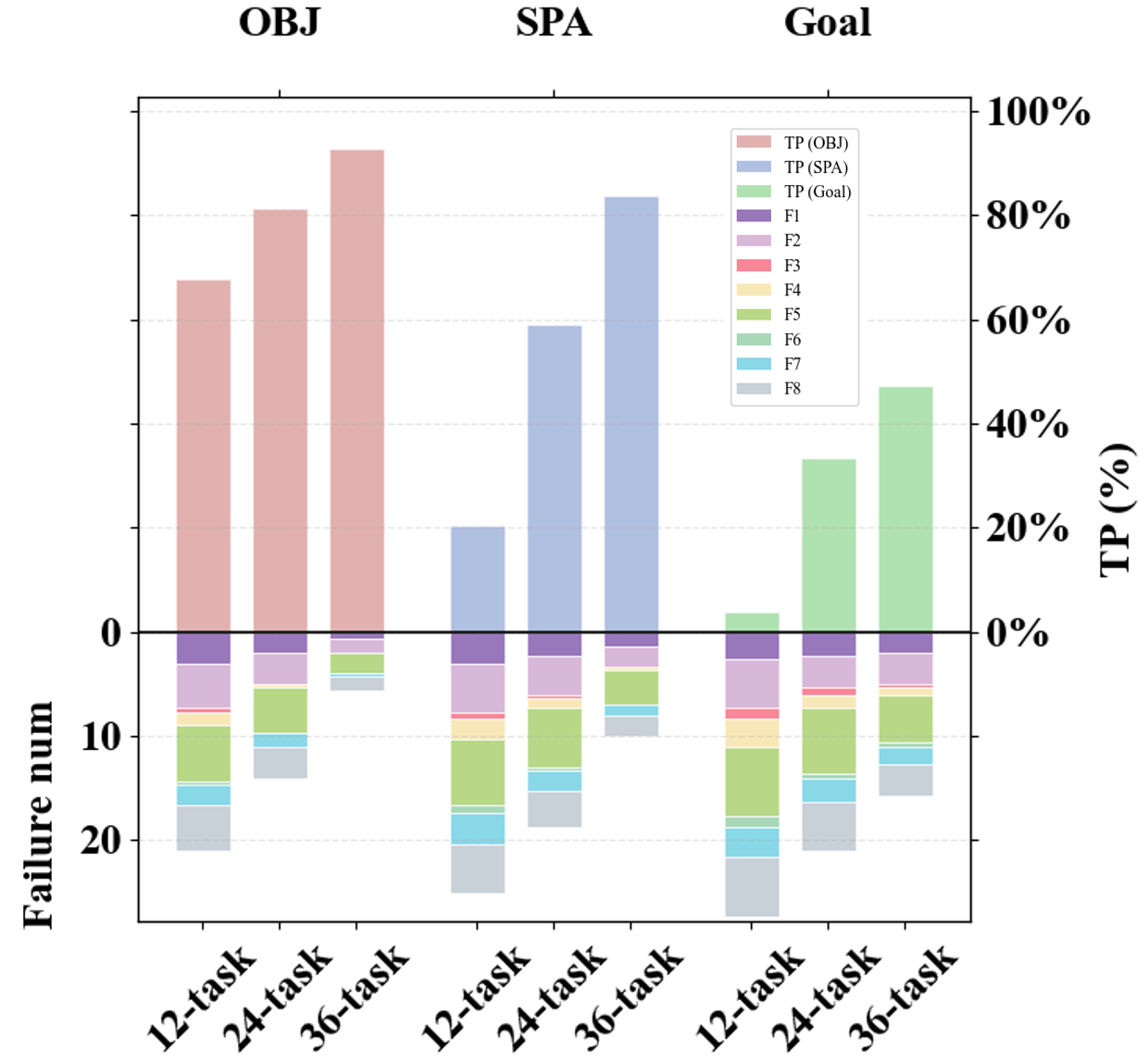}
    \caption{Ablation results of data scale}
    \label{fig:ablation_scale}
\end{figure}

\textbf{Q1: Does simply scaling up task-specific data solve complex physical reasoning?}
Figure.~\ref{fig:ablation_scale} shows the effect of scaling task-specific trajectories from 12 to 36. 
While TP on Object tasks improves rapidly and saturates with increased data—primarily due to reductions in HE and GUE—the gains on SAS and GCG tasks remain limited. 
Even when the dataset size is doubled, improvements on GCG tasks are limited, with PE and CSR decreasing by only 3--5\%. 
This trend indicates that homogeneous data scaling is effective for resolving object-level ambiguities but is insufficient for addressing long-horizon causal reasoning and spatial planning, which dominate failures in SAS and GCG tasks.

\textbf{Q2: How do different components of the Manip-Cognition dataset contribute to model capabilities?}

\begin{figure}[ht]
    \centering
    \includegraphics[width=1\linewidth]{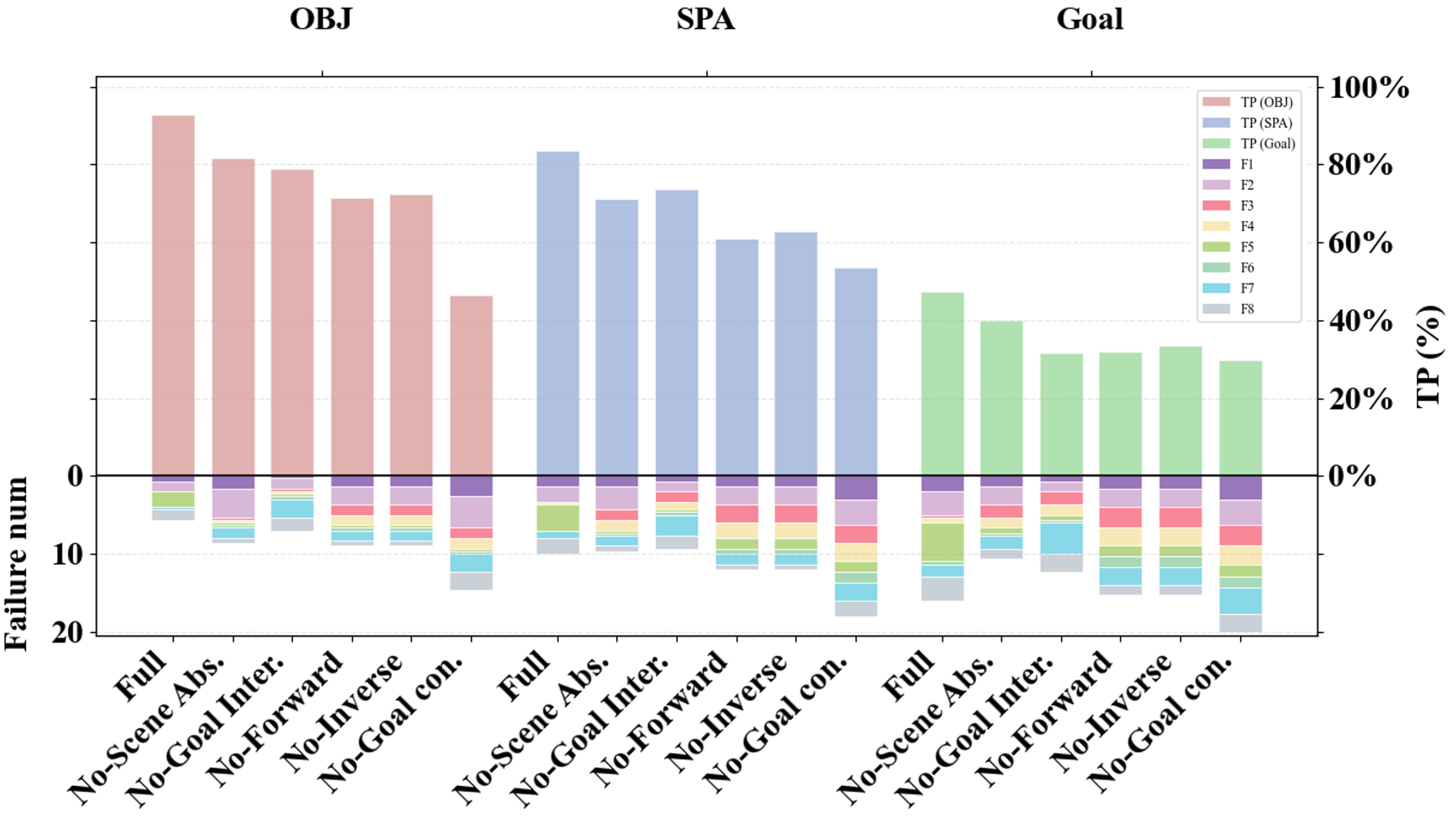}
    \caption{Ablation results of data compositions}
    \label{fig:ablation_composition}
\end{figure}

Figure.~\ref{fig:ablation_composition} illustrates the contribution of individual data components. 
Removing Action Planning Data, especially goal-conditioned planning trajectories, leads to the most severe performance degradation, reducing TP by up to 50\% across all task categories and sharply increasing both HE and GUE. 
This highlights the central role of explicit planning supervision in long-horizon reasoning.
Removing Scene Abstraction Data disproportionately affects SAS tasks, correlating with higher object-level hallucinations, as reflected by increased HE.
In contrast, removing Forward Scene Prediction or Inverse Action Reasoning mainly increases PE, indicating their importance in learning action–state causality.
Finally, excluding Goal Interpretation Data causes a pronounced rise in GUE, confirming its necessity for accurate task termination and goal satisfaction.

\textbf{Q3: How does data augmentation and noise injection affect generalization and robustness?}

\begin{figure}[ht]
    \centering
    \includegraphics[width=0.8\linewidth]{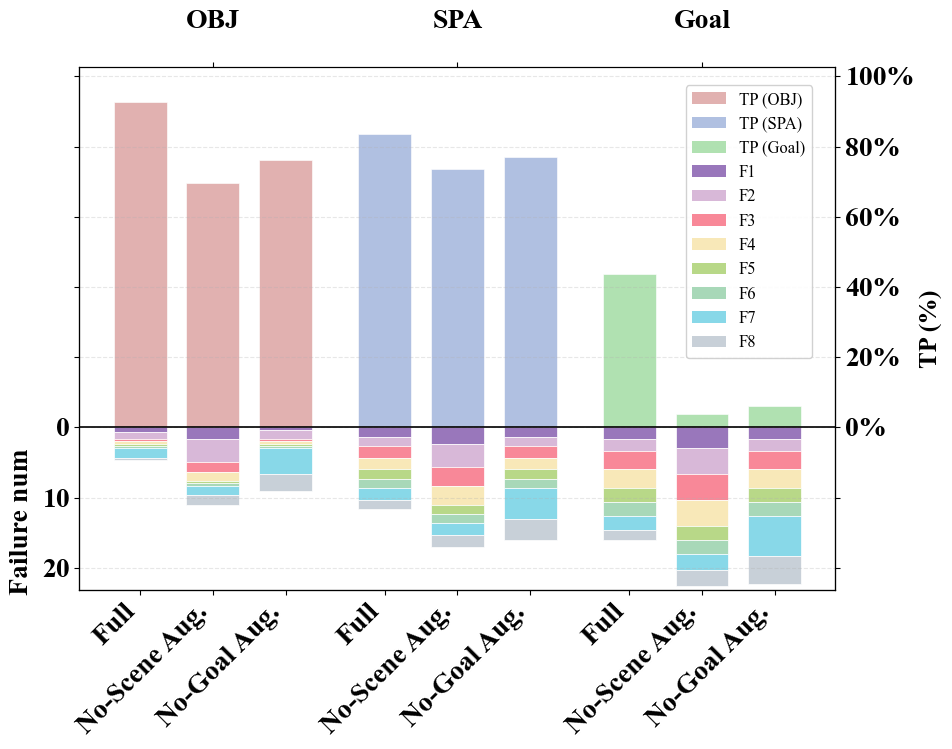}
    \caption{Ablation results on data augmentation methods}
    \label{fig:ablation_augmentation}
\end{figure}

The impact of data augmentation is summarized in Figure.~\ref{fig:ablation_augmentation}.

\begin{figure}[ht]
    \centering
    \includegraphics[width=0.8\linewidth]{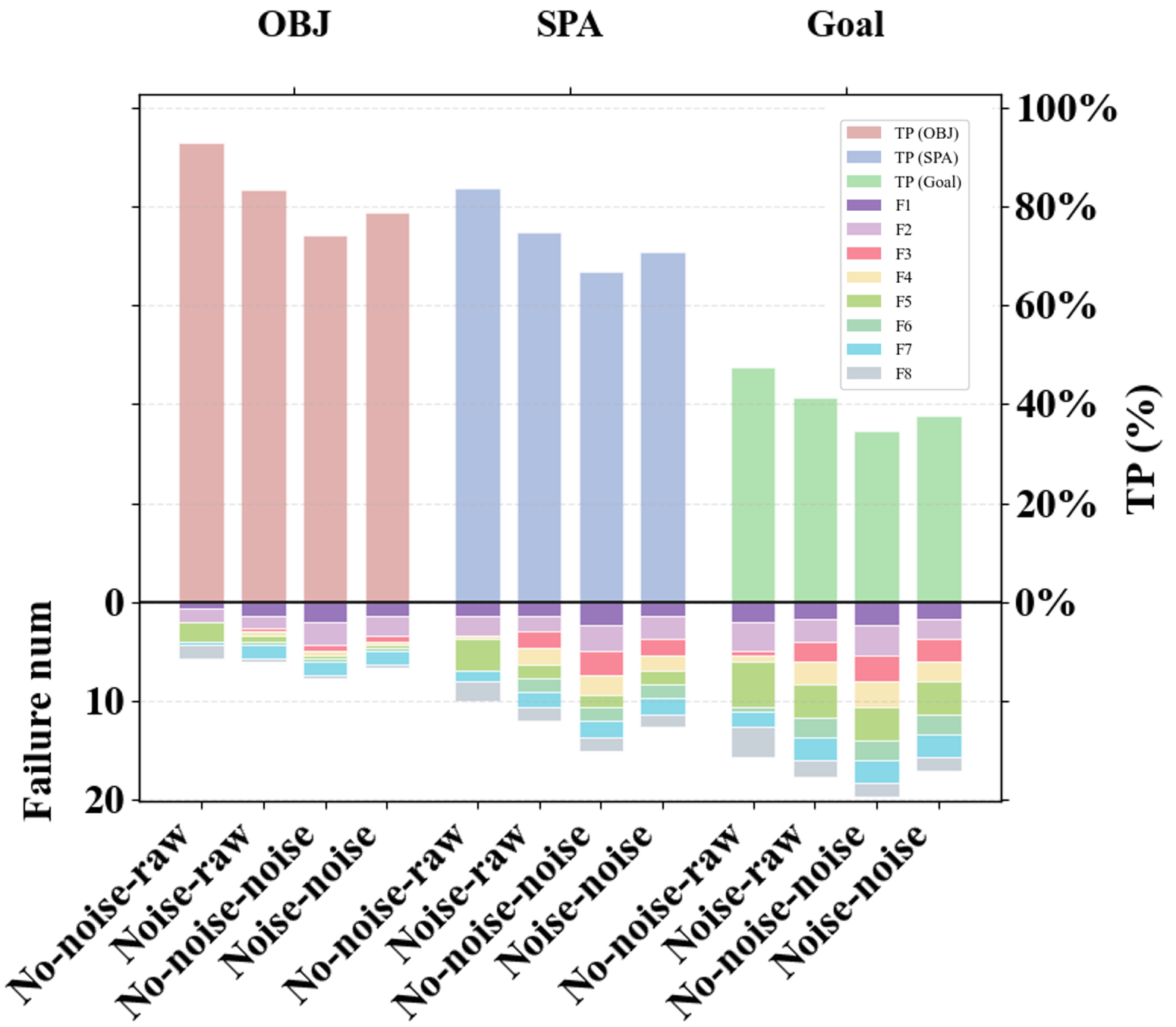}
    \caption{Efficacy of Denoised Training: Comparing Model Performance Under Clean and Noisy Scene Graph Conditions}
    \label{fig:ablation_noise_robustness}
\end{figure}

Scene-level augmentation further improves spatial reasoning and action inference, leading to consistent reductions in PE, particularly on SPA tasks.
Similarly, goal augmentation enhances robustness in task completion, effectively suppressing GUE such as premature termination.
Figure.~\ref{fig:ablation_noise_robustness} illustrates that training with noisy scene graphs introduces a small drop in TP under clean evaluation conditions, mainly due to a slight increase in HE.
However, when evaluated under corrupted or incomplete scene graphs, models trained with noise injection exhibit substantially more stable performance and stronger resistance to perceptual errors.


\textbf{Q4: How does Reinforcement Fine-Tuning improve complex reasoning compared to Standard Supervised Fine-Tuning?}

\begin{figure}[ht]
    \centering
    \includegraphics[width=0.8\linewidth]{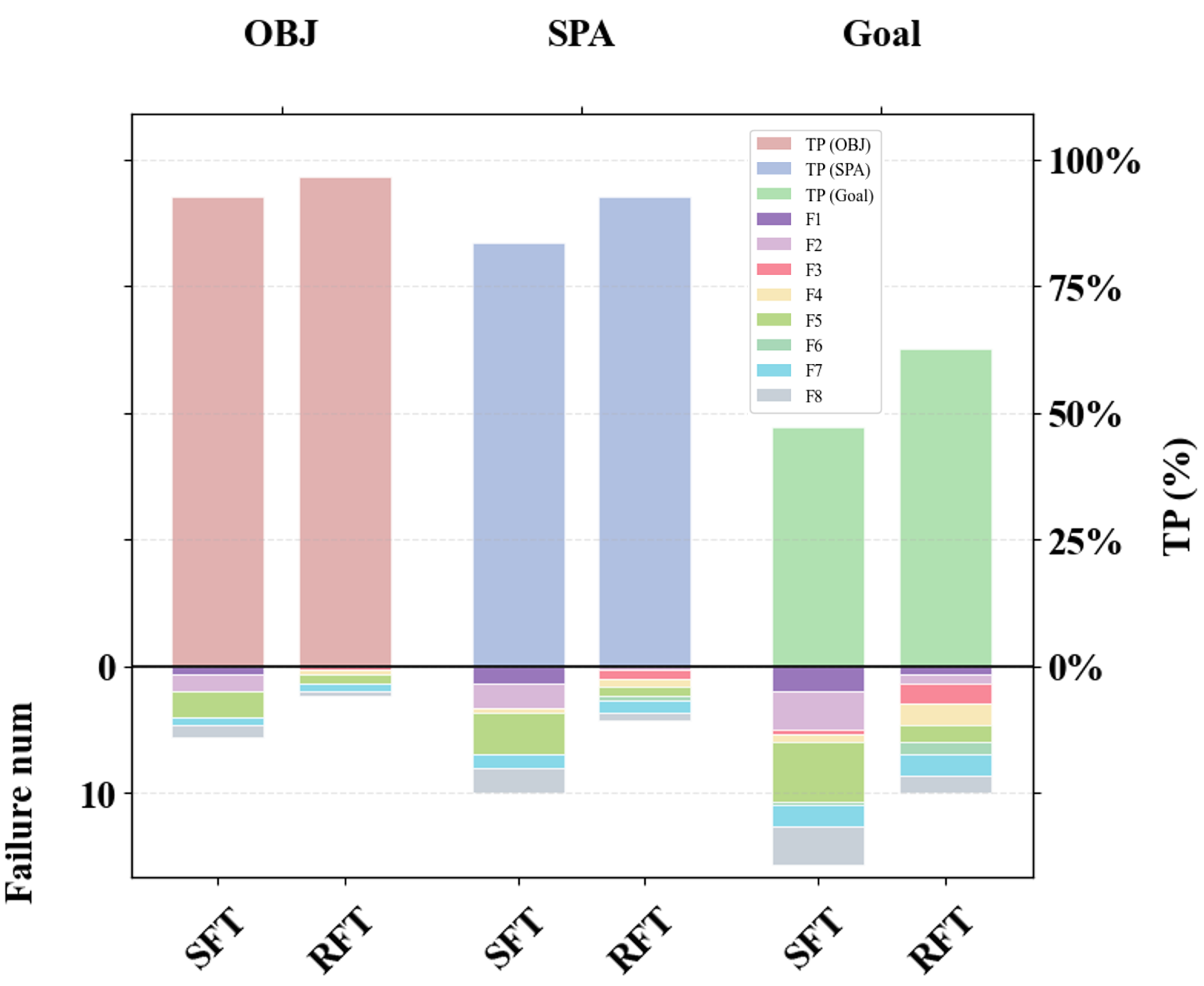}
    \caption{Ablation results of training paradigms}
    \label{fig:ablation_training_paradigm}
\end{figure}

Figure.~\ref{fig:ablation_training_paradigm} compares SFT and RFT across all task categories.
Applying RFT on top of SFT consistently improves TP, with an 11.7\% absolute gain on the most challenging Spatial tasks.
This improvement is directly reflected in the failure analysis.
HE are substantially reduced due to the \textit{Scene-Graph Grounding Reward} ($R_G$), which penalizes predictions of non-existent objects, and the \textit{Step-wise Constraint Reward} ($R_S$), which prevents multi-action hallucinations.
Additionally, CSR errors decrease as $R_S$ enforces strict single-step, closed-loop execution, discouraging physically implausible continuous actions.
Finally, GUE, such as premature stopping, are mitigated by the \textit{Termination Reward} ($R_T$), which penalizes generating the \texttt{END} token before the goal condition is satisfied.
These results confirm that RFT is critical for aligning reasoning trajectories with physical and goal constraints beyond what supervised learning alone can achieve.

\subsection{Demonstration in Real-World Tasks}
\label{section: realworld evaluation}

We validate RoBoSR through real-world experiments designed to probe its three core capabilities. Each experiment targets a specific aspect of structured reasoning and action execution, as demonstrated in the supplementary videos.
For instruction-driven long-horizon tasks, such as \textit{``Making a coffee''}, \textbf{we collect 50 ego-centric VR demonstrations ffor each of tasks}. The dataset includes both successful executions and failure cases with corresponding recovery actions.

\noindent\textbf{1) Instruction–State Alignment:}
We evaluate the ability to map high-level instructions to scene-consistent actions using language-guided pick-and-place tasks. Examples include \textit{"Pick up the red mug and place it on the table''} and \textit{''Move the blue block next to the green cube''}.

\noindent\textbf{2) Causally Constrained State-Space Reasoning:}
To assess long-horizon planning and causal subtask reasoning, we test multi-step daily-life tasks such as \textit{''Object sorting''}, and \textit{''Prepare a simple breakfast''} and \textit{''Set the table with plates and cups in order''}. RoBoSR maintains an state representation and enforces causal constraints between subtasks.

\noindent\textbf{3) State-Consistency Correction:}
We evaluate robustness to execution deviations by introducing perturbations, e.g., objects are displaced or dropped during task execution. Tasks include table organization or item placement into containers. RoBoSR detects inconsistencies between the observed and expected states and autonomously corrects errors. 

\begin{figure*}[ht]
    \centering
    \includegraphics[width=1\linewidth]{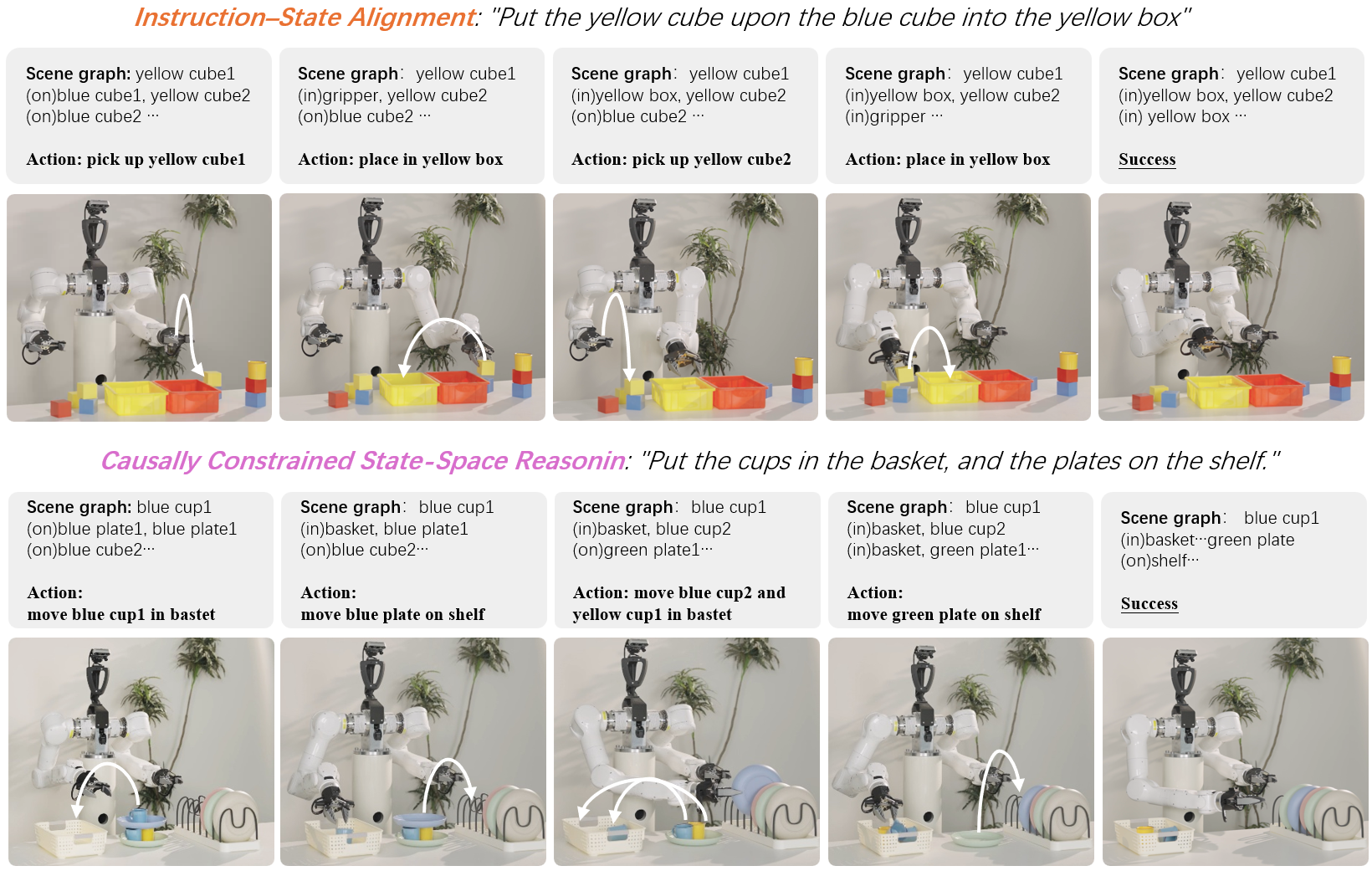}
    \caption{Real-world demonstrations of RoBoSR on instruction-driven manipulation tasks.
\textbf{Top:} Instruction–State Alignment for a language-guided task (\textit{``Put the yellow cube upon the blue cube into the yellow box''}), where RoBoSR maps high-level instructions to scene-consistent actions.
\textbf{Bottom:} Causally Constrained State-Space Reasoning for a long-horizon task (\textit{``Put the cups in the basket, and the plates on the shelf''}), demonstrating structured subtask sequencing under causal and spatial constraints.}
    \label{fig:Demonstration in Real World Tasks}
\end{figure*}

\section{Conclusion} \label{sec: conclusion}
We presented RoBoSR, a representation-centric framework for embodied manipulation that reasons over explicit world models rather than direct perception-to-action mappings. By explicitly modeling object states and their relations, RoBoSR enables structured and compositional reasoning over evolving environments. This structural grounding supports principled generalization, coherent long-horizon planning, and transferable reasoning capabilities, as demonstrated in both simulated and real-world experiments.

\noindent\textbf{Limitations and Future Works:}
Despite its advantages, our approach has several certain limitations.
First, RoBoSR is limited in representing highly deformable objects and fine-grained continuous states (e.g., partial opening angles of articulated objects). This limitation arises from the reliance on discretized symbolic states, which cannot fully capture high-dimensional continuous variations. Future work will extend RoBoSR toward continuous and deformable state representations to enable finer-grained reasoning and control.
Second, RoBoSR depends on a strong perception front-end to construct structured scene graphs from raw sensory inputs. This requirement stems from the need for accurate object detection, relation inference, and state estimation to support downstream reasoning. Future research will investigate more robust and scalable perception-to-structure grounding, including tighter perception–reasoning integration and uncertainty-aware state estimation.

\bibliographystyle{IEEEtran}
\bibliography{bibtex/bib/main}

@inproceedings{gu2024conceptgraphs,
  title={Conceptgraphs: Open-vocabulary 3d scene graphs for perception and planning},
  author={Gu, Qiao and Kuwajerwala, Ali and Morin, Sacha and Jatavallabhula, Krishna Murthy and Sen, Bipasha and Agarwal, Aditya and Rivera, Corban and Paul, William and Ellis, Kirsty and Chellappa, Rama and others},
  booktitle={2024 IEEE International Conference on Robotics and Automation (ICRA)},
  pages={5021--5028},
  year={2024},
  organization={IEEE}
}

@article{huang2023voxposer,
  title={Voxposer: Composable 3d value maps for robotic manipulation with language models},
  author={Huang, Wenlong and Wang, Chen and Zhang, Ruohan and Li, Yunzhu and Wu, Jiajun and Fei-Fei, Li},
  journal={arXiv preprint arXiv:2307.05973},
  year={2023}
}

@article{huang2024rekep,
  title={Rekep: Spatio-temporal reasoning of relational keypoint constraints for robotic manipulation},
  author={Huang, Wenlong and Wang, Chen and Li, Yunzhu and Zhang, Ruohan and Fei-Fei, Li},
  journal={arXiv preprint arXiv:2409.01652},
  year={2024}
}

@article{ai2025review,
  title={A review of learning-based dynamics models for robotic manipulation},
  author={Ai, Bo and Tian, Stephen and Shi, Haochen and Wang, Yixuan and Pfaff, Tobias and Tan, Cheston and Christensen, Henrik I and Su, Hao and Wu, Jiajun and Li, Yunzhu},
  journal={Science Robotics},
  volume={10},
  number={106},
  pages={eadt1497},
  year={2025},
  publisher={American Association for the Advancement of Science}
}

@article{wang2025enact,
  title={ENACT: Evaluating Embodied Cognition with World Modeling of Egocentric Interaction},
  author={Wang, Qineng and Huang, Wenlong and Zhou, Yu and Yin, Hang and Bao, Tianwei and Lyu, Jianwen and Liu, Weiyu and Zhang, Ruohan and Wu, Jiajun and Fei-Fei, Li and others},
  journal={arXiv preprint arXiv:2511.20937},
  year={2025}
}

@article{chen2023egoplan,
  title={Egoplan-bench: Benchmarking multimodal large language models for human-level planning},
  author={Chen, Yi and Ge, Yuying and Ge, Yixiao and Ding, Mingyu and Li, Bohao and Wang, Rui and Xu, Ruifeng and Shan, Ying and Liu, Xihui},
  journal={arXiv preprint arXiv:2312.06722},
  year={2023}
}

@article{chen2024roboscript,
  title={Roboscript: Code generation for free-form manipulation tasks across real and simulation},
  author={Chen, Junting and Mu, Yao and Yu, Qiaojun and Wei, Tianming and Wu, Silang and Yuan, Zhecheng and Liang, Zhixuan and Yang, Chao and Zhang, Kaipeng and Shao, Wenqi and others},
  journal={arXiv preprint arXiv:2402.14623},
  year={2024}
}

@article{belkhale2024rt,
  title={Rt-h: Action hierarchies using language},
  author={Belkhale, Suneel and Ding, Tianli and Xiao, Ted and Sermanet, Pierre and Vuong, Quon and Tompson, Jonathan and Chebotar, Yevgen and Dwibedi, Debidatta and Sadigh, Dorsa},
  journal={arXiv preprint arXiv:2403.01823},
  year={2024}
}

@inproceedings{zhao2025cot,
  title={Cot-vla: Visual chain-of-thought reasoning for vision-language-action models},
  author={Zhao, Qingqing and Lu, Yao and Kim, Moo Jin and Fu, Zipeng and Zhang, Zhuoyang and Wu, Yecheng and Li, Zhaoshuo and Ma, Qianli and Han, Song and Finn, Chelsea and others},
  booktitle={Proceedings of the Computer Vision and Pattern Recognition Conference},
  pages={1702--1713},
  year={2025}
}

@article{black2024pi_0,
  title={PI0: A Vision-Language-Action Flow Model for General Robot Control},
  author={Black, Kevin and Brown, Noah and Driess, Danny and Esmail, Adnan and Equi, Michael and Finn, Chelsea and Fusai, Niccolo and Groom, Lachy and Hausman, Karol and Ichter, Brian and others},
  journal={arXiv preprint arXiv:2410.24164},
  year={2024}
}

@article{li2024behavior,
  title={Behavior-1k: A human-centered, embodied ai benchmark with 1,000 everyday activities and realistic simulation},
  author={Li, Chengshu and Zhang, Ruohan and Wong, Josiah and Gokmen, Cem and Srivastava, Sanjana and Mart{\'\i}n-Mart{\'\i}n, Roberto and Wang, Chen and Levine, Gabrael and Ai, Wensi and Martinez, Benjamin and others},
  journal={arXiv preprint arXiv:2403.09227},
  year={2024}
}

@article{damen2020epic,
  title={The epic-kitchens dataset: Collection, challenges and baselines},
  author={Damen, Dima and Doughty, Hazel and Farinella, Giovanni Maria and Fidler, Sanja and Furnari, Antonino and Kazakos, Evangelos and Moltisanti, Davide and Munro, Jonathan and Perrett, Toby and Price, Will and others},
  journal={IEEE Transactions on Pattern Analysis and Machine Intelligence},
  volume={43},
  number={11},
  pages={4125--4141},
  year={2020},
  publisher={IEEE}
}

@inproceedings{ji2025robobrain,
  title={Robobrain: A unified brain model for robotic manipulation from abstract to concrete},
  author={Ji, Yuheng and Tan, Huajie and Shi, Jiayu and Hao, Xiaoshuai and Zhang, Yuan and Zhang, Hengyuan and Wang, Pengwei and Zhao, Mengdi and Mu, Yao and An, Pengju and others},
  booktitle={Proceedings of the Computer Vision and Pattern Recognition Conference},
  pages={1724--1734},
  year={2025}
}

@article{kim2024openvla,
  title={Openvla: An open-source vision-language-action model},
  author={Kim, Moo Jin and Pertsch, Karl and Karamcheti, Siddharth and Xiao, Ted and Balakrishna, Ashwin and Nair, Suraj and Rafailov, Rafael and Foster, Ethan and Lam, Grace and Sanketi, Pannag and others},
  journal={arXiv preprint arXiv:2406.09246},
  year={2024}
}

@article{yang2023foundation,
  title={Foundation models for decision making: Problems, methods, and opportunities},
  author={Yang, Sherry and Nachum, Ofir and Du, Yilun and Wei, Jason and Abbeel, Pieter and Schuurmans, Dale},
  journal={arXiv preprint arXiv:2303.04129},
  year={2023}
}

@article{li2024scene,
  title={Scene graph generation: A comprehensive survey},
  author={Li, Hongsheng and Zhu, Guangming and Zhang, Liang and Jiang, Youliang and Dang, Yixuan and Hou, Haoran and Shen, Peiyi and Zhao, Xia and Shah, Syed Afaq Ali and Bennamoun, Mohammed},
  journal={Neurocomputing},
  volume={566},
  pages={127052},
  year={2024},
  publisher={Elsevier}
}

@article{brohan2022rt,
  title={Rt-1: Robotics transformer for real-world control at scale},
  author={Brohan, Anthony and Brown, Noah and Carbajal, Justice and Chebotar, Yevgen and Dabis, Joseph and Finn, Chelsea and Gopalakrishnan, Keerthana and Hausman, Karol and Herzog, Alex and Hsu, Jasmine and others},
  journal={arXiv preprint arXiv:2212.06817},
  year={2022}
}

@inproceedings{zitkovich2023rt,
  title={Rt-2: Vision-language-action models transfer web knowledge to robotic control},
  author={Zitkovich, Brianna and Yu, Tianhe and Xu, Sichun and Xu, Peng and Xiao, Ted and Xia, Fei and Wu, Jialin and Wohlhart, Paul and Welker, Stefan and Wahid, Ayzaan and others},
  booktitle={Conference on Robot Learning},
  pages={2165--2183},
  year={2023},
  organization={PMLR}
}

@misc{qwen8B_system_card_2025,
      author={{Qwen Team}},
      title={{Qwen3: Think Deeper, Act Faster}},
      year={2025},
      month= apr,
      url={https://openai.com/zh-Hans-CN/index/gpt-5-system-card/},
      note={Accessed: 2025-04-29}
}

@misc{GSRbench_2026,
      author={{GSRbench}},
      title={{Ground scene reasoning(GSR)-bench}},
      year={2026},
      month= apr,
      url={https://github.com/KLMmotion/GSR-bench/},
}

@article{wang2025skil,
  title={Skil: Semantic keypoint imitation learning for generalizable data-efficient manipulation},
  author={Wang, Shengjie and You, Jiacheng and Hu, Yihang and Li, Jiongye and Gao, Yang},
  journal={arXiv preprint arXiv:2501.14400},
  year={2025}
}




\end{document}